\documentclass{article}

\usepackage{amssymb,amsthm,mathtools}
\usepackage{arxiv}
\usepackage{amsmath}
\usepackage[utf8]{inputenc} 
\usepackage[T1]{fontenc}    
\usepackage{hyperref}       
\usepackage{url}            
\usepackage{booktabs}       
\usepackage{amsfonts}       
\usepackage{nicefrac}       
\usepackage{microtype}      
\usepackage{lipsum}
\usepackage{graphicx}
\graphicspath{ {./images/} }

\newtheorem{remark}{Remark}

\title{A Honest Cross-Validation Estimator for Prediction Performance}
\author{
Tianyu Pan \\
Department of Biomedical Data Science\\ 
Stanford University\\
\and 
Vincent Z. Yu\\
Indiana Academy for Science, Mathematics, and Humanities\\
\and
Viswanath Devanarayan \\
Eisai Inc.\\ 
\and 
Lu Tian \\ Department of Biomedical Data Science \\ Stanford University\\ lutian@stanford.edu}
\date{\today}

\begin{document}
\maketitle

\begin{abstract}
Cross-validation is a standard tool for obtaining a honest assessment of the performance of a prediction model. The commonly used version repeatedly splits data, trains the prediction model on the training set, evaluates the model performance on the test set, and averages the model performance across different data splits. A well-known criticism is that such cross-validation procedure does not directly estimate the performance of the particular model recommended for future use. In this paper, we propose a new method to estimate the performance of a model trained on a specific (random) training set. A naive estimator can be obtained by applying the model to a disjoint testing set. Surprisingly, cross-validation estimators computed from other random splits can be used to improve this naive estimator within a random-effects model framework. We develop two estimators---a hierarchical Bayesian estimator and an empirical Bayes estimator---that perform similarly to or better than both the conventional cross-validation estimator and the naive single-split estimator. Simulations and a real-data example demonstrate the superior performance of the proposed method.
\end{abstract}

\section{Introduction}
Cross-validation is widely used to evaluate predictive models, but its target parameter is often misunderstood. Importantly, cross-validation does not estimate the performance of a single fitted model from a fixed training set; rather, it targets the average performance of a collection of models, each trained on a different partition of the data. Thus, cross-validation evaluates a \emph{modeling strategy} across repeated samples, not the accuracy of one specific model realization \cite{bates2024cross, bayle2020cross, cai2023bootstrapping, efron1997improvements}.

Among its many forms, $K$-fold cross-validation is perhaps the most popular. Conceptually, it can be viewed as a computationally efficient, balanced approximation to an idealized cross-validation procedure in which the data are randomly split into training and testing sets infinitely many times. By dividing the data into $K$ folds and cycling each fold as a test set, $K$-fold cross-validation approximates this ``infinite-split'' average with smaller variance while ensuring that every observation contributes both to model training and to performance evaluation.  But conceptually, the $K$-fold cross-validation still evaluate a \emph{modeling strategy} rather than a single model.

A simpler alternative in practice is a \emph{single split}: train the model on a designated training set and evaluate its performance on the held-out testing set. This yields a valid estimate of predictive accuracy for that particular train--test split and allows one to recommend a specific model with an accompanying performance estimate to users. However, this strategy may suffer from high variability: the estimate can depend strongly on the arbitrary split, the test set may be small, and practitioners might be tempted to report only favorable splits, introducing a risk of cherry-picking.

We follow the single-split paradigm but use additional information to improve the estimate of the performance of the model trained on a training set. Somewhat unexpectedly, the conventional cross-validation estimates---although targeting evaluating modeling strategy---are still informative for the performance of this specific model. We formalize this using a mixed effects model framework in which repeated cross-validation helps refine the estimate based on a single held-out testing set.

The paper is organized as follows. Section 2 introduces the general theoretical framework and illustrates it with two specific applications. Section 3 presents the simulation results, Section 4 demonstrates a real data analysis using the proposed methods, and Section 5 concludes with a brief discussion of potential extensions.

\section{Method}
\subsection{Estimation}
Suppose that we observe $n$ independent identically distributed (i.i.d) observations $\mathbf{X}=\{X_1,\ldots,X_n\}$ and randomly split the data into two disjoint parts
\[
\mathbf{X}=\mathbf{D}_1^{(0)}\cup\mathbf{D}_2^{(0)},
\]
with sizes $n_1$ and $n_2$ ($n_1+n_2=n$), respectively. We may train a prediction model on $\mathbf{D}_1^{(0)}$, denoted by $\widehat{\mathcal{M}}_{\mathbf{D}_1^{(0)}}$, and evaluate its performance on $\mathbf{D}_2^{(0)}$ via a loss function
\[
\widehat{\mathrm{Err}}_0=\frac{1}{n_2}\sum_{X_i\in\mathbf{D}_2^{(0)}}L\bigl(X_i,\widehat{\mathcal{M}}_{\mathbf{D}_1^{(0)}}\bigr).
\]
Here $L(X,\mathcal{M})$ quantifies predictive loss when applying model $\mathcal{M}$ to observation $X$. One may expect that $\widehat{\mathrm{Err}}_0$ consistently estimates
\[
\mathrm{Err}_0=\mathbb{E}\bigl\{L\bigl(X_0,\widehat{\mathcal{M}}_{\mathbf{D}_1^{(0)}}\bigr)\bigr\},
\]
where the expectation is with respect to an independent draw $X_0.$  Here $\mathrm{Err}_0$ is the parameter of our interest, as it is useful for future users of the model $\widehat{{\cal M}}_{\mathbf{D}_1^{(0)}}.$ Under mild conditions, conditional on $\widehat{\mathcal{M}}_{\mathbf{D}_1^{(0)}}$, a normal approximation holds for moderately large $n_2$:
\[
\widehat{\mathrm{Err}}_0\mid \widehat{\mathcal{M}}_{\mathbf{D}_1^{(0)}}\;\sim \;\mathcal{N}\bigl(\mathrm{Err}_0,\,\widehat{\sigma}_0^{2}\bigr),\qquad
\widehat{\sigma}_0^{2}=\frac{1}{n_2^2}\sum_{X_i\in\mathbf{D}_2^{(0)}}\Bigl\{L\bigl(X_i,\widehat{\mathcal{M}}_{\mathbf{D}_1^{(0)}}\bigr)-\widehat{\mathrm{Err}}_0\Bigr\}^2.
\]
In summary, $\widehat{\mathrm{Err}}_0$ can be viewed as a natural (naive) estimator of $\mathrm{Err}_0$.

There are many random splits into training and testing sets of the size of $n_1$ and $n_2$, respectively. Indexing the $k$th split by a superscript, write
\[
\mathbf{X}=\mathbf{D}_1^{(k)}\cup\mathbf{D}_2^{(k)},\quad k=1,\ldots,K.
\]
Let the true prediction performance for the model trained on $\mathbf{D}_1^{(k)}$ be
\[
\mathrm{Err}_k=\mathbb{E}\bigl\{L\bigl(X_0,\widehat{\mathcal{M}}_{\mathbf{D}_1^{(k)}}\bigr)\bigr\},
\]
which can be estimated by
\[
\widehat{\mathrm{Err}}_k=\frac{1}{n_2}\sum_{X_i\in\mathbf{D}_2^{(k)}}L\bigl(X_i,\widehat{\mathcal{M}}_{\mathbf{D}_1^{(k)}}\bigr),\]
where
\[
\widehat{\mathrm{Err}}_k\mid \widehat{\mathcal{M}}_{\mathbf{D}_1^{(k)}}\;\sim\;\mathcal{N}\bigl(\mathrm{Err}_k,\,\widehat{{\sigma}}_k^{2}\bigr), \qquad
\widehat{{\sigma}}_k^{2}=\frac{1}{n_2^2}\sum_{X_i\in\mathbf{D}_2^{(k)}}\Bigl\{L\bigl(X_i,\widehat{\mathcal{M}}_{\mathbf{D}_1^{(k)}}\bigr)-\widehat{\mathrm{Err}}_k\Bigr\}^2.
\]

Now, we can describe our statistical model. First, we assume that models trained on $\mathbf{D}_1^{(0)},\ldots,\mathbf{D}_1^{(K)}$ are comparable so that their true performances satisfy a random-effects model as in meta analysis \cite{michael2019exact}
\begin{equation}\label{eq:randomeffect}
\mathrm{Err}_k\stackrel{\text{iid}}{\sim}\mathcal{N}(\mu_0,\tau_0^2),\qquad k=0,1,\ldots,K.
\end{equation}
Second, we have a model for the conditional distribution of their empirical estimates:
\begin{equation} \label{eq:randomerror}
\begin{pmatrix}\widehat{\mathrm{Err}}_0\\ \widehat{\mathrm{Err}}_1\\ \vdots\\ \widehat{\mathrm{Err}}_K\end{pmatrix} \biggr | \begin{pmatrix}\mathrm{Err}_0\\ \mathrm{Err}_1\\ \vdots\\ \mathrm{Err}_K\end{pmatrix}
\sim \mathcal{N}\!\left(\begin{pmatrix}\mathrm{Err}_0\\ \mathrm{Err}_1\\ \vdots\\ \mathrm{Err}_K\end{pmatrix},\;\widehat{\boldsymbol{\Sigma}}\right),
\end{equation}
where $\widehat{\boldsymbol{\Sigma}}$ can be estimated from the data and is treated as known. Note that $\mathrm{Cov}(\widehat{\mathrm{Err}}_k,\widehat{\mathrm{Err}}_\ell)$ is generally nonzero, if $\mathbf{D}_2^{(k)}\cap\mathbf{D}_2^{(\ell)}\neq\varnothing$. For most loss functions used in practice these covariances can be estimated analytically (see below). (\ref{eq:randomeffect}) and (\ref{eq:randomerror}) together form a random effects model.

Our goal is to estimate $\mathrm{Err}_0$, the performance of $\widehat{\mathcal{M}}_{\mathbf{D}_1^{(0)}}$, based on $\{\widehat{\mathrm{Err}}_k\}_{k=0}^K$ and $\widehat{\boldsymbol{\Sigma}}.$ Under the random effects model specified in (\ref{eq:randomeffect}) and (\ref{eq:randomerror}), there are multiple ways to achieve this objective.  First, we consider a hierarchical Bayesian approach by adopting a weakly-informative prior
\[
\tau_0^{-2}\sim\mathrm{Gamma}(a_0,b_0),\qquad \mu_0\mid \tau_0^2\sim \mathcal{N}\!\bigl(0,\,\tau_0^2/\kappa_0\bigr),
\]
with small $a_0,b_0,\kappa_0>0$. This approach allows to sample $\mathrm{Err}_0$ from its posterior distribution.  To this end, a Gibbs sampler alternates between the following:
\begin{enumerate}
  \item Given $(\mu_0,\tau_0^2)$,
  \begin{equation}
  \label{eq:gibb1}
  \mathbf{Err} \Big| \widehat{\mathbf{Err}},\mu_0,\tau_0^2
  \sim \mathcal{N}\!\left(\mu_0\mathbf{1}+\mathbf{B}\bigl(\widehat{\mathbf{Err}}-\mu_0\mathbf{1}\bigr),\; (\mathbf{I}_{K+1}-\mathbf{B})\,\widehat{\boldsymbol{\Sigma}}\right),
  \end{equation}
  where $\mathbf{Err}=(\mathrm{Err}_0, \dots, \mathrm{Err}_K)^\top$,  $\widehat{\mathbf{Err}}=(\widehat{\mathrm{Err}}_0,\ldots,\widehat{\mathrm{Err}}_K)^\top$, $\mathbf{1}$ is the $(K{+}1)$-vector of ones, and
  \[
  \mathbf{B}=\tau_0^2\bigl(\tau_0^2\mathbf{I}_{K+1}+\widehat{\boldsymbol{\Sigma}}\bigr)^{-1}.
  \]
  \item Given $\mathbf{Err},$
  \begin{equation}
  \label{eq:gibb2}
  \tau_0^{-2}\mid \mathbf{Err} \sim\mathrm{Gamma}\!\left(a_1,b_1\right),\qquad \mu_0\mid \tau_0^2, \mathbf{Err} \sim \mathcal{N}\!\left(\mu_1,\,\frac{\tau_0^2}{\kappa_1}\right),
  \end{equation}
  where 
  $$a_1=a_0+\tfrac{K+1}{2},~ \kappa_1=\kappa_0+K+1,~ \bar{\mathrm{Err}}=\frac{1}{K+1}\sum_{k=0}^K\mathrm{Err}_k, ~\mu_1=\frac{\kappa_0\cdot 0+(K+1)\bar{\mathrm{Err}}}{\kappa_1},$$ and $$b_1=b_0+\frac{1}{2}\sum_{k=0}^K(\mathrm{Err}_k-\bar{\mathrm{Err}})^2+\frac{\kappa_0(K+1)}{2\kappa_1}\,\bar{\mathrm{Err}}^{\,2}.$$
\end{enumerate}
Iterating \eqref{eq:gibb1}--\eqref{eq:gibb2} yields a Markov chain $\{\mathrm{Err}_0^{[m]}\}_{m=1}^M$ for the parameter of interest. The corresponding Bayesian point estimator of $\mathrm{Err}_0$ is
\begin{equation}
\label{eq:Best}
\widehat{\mathrm{Err}}_0^{B}=\frac{1}{M-M_0}\sum_{m=M_0+1}^M \mathrm{Err}_0^{[m]},
\end{equation}
with the first $M_0$ draws discarded as burn-in. Because it borrows strength from $\{\widehat{\mathrm{Err}}_k\}_{k=0}^K$, $\widehat{\mathrm{Err}}_0^{B}$ is typically more accurate than the naive $\widehat{\mathrm{Err}}_0$.

\begin{remark}
By the exchangeability of $\{\widehat{\mathrm{Err}}_k\}_{k=0}^K$, a parsimonious choice for $\widehat{\boldsymbol{\Sigma}}$ is its compound-symmetric form with diagonal elements being
\(
(K+1)^{-1}\sum_{k=0}^K\widehat{\sigma}_k^2
\)
and common off-diagonal elements being
\(
2\{K(K+1)\}^{-1}\sum_{0\le i<j\le K}\widehat{\sigma}_{i,j}.
\)
This version of $\widehat{\boldsymbol{\Sigma}}$ stabilizes the inverse operation involving matrix $\widehat{\boldsymbol{\Sigma}}$, especially when the dimension $K+1$ is not small. Therefore, we recommend use this compound-symmetric version of $\widehat{\boldsymbol{\Sigma}}$ in hierarchical Bayesian procedure. 
\end{remark}

\begin{remark}
In addition to a point estimate, a $(1-\alpha)$ credible interval for $\mathrm{Err}_0$ can be obtained as the $(\alpha/2)\times 100$ and $(1-\alpha/2)\times 100$th percentiles of $\{\mathrm{Err}_0^{[m]}\}_{m=M_0+1}^M$. Although derived under a Bayesian model, such intervals often exhibit approximate nominal frequentist coverage in our numerical experiments in section \ref{sec:simulation}.
\end{remark}

Alternatively, a simple empirical Bayes (EB) plug-in estimator for $\mathrm{Err}_0$ can be obtained by first estimating $\mu_0$ and $\tau_0^2$ by moment estimates
\begin{equation} \label{eq:momentest}
\widehat{\mu}_0=\frac{1}{K+1}\sum_{k=0}^K \widehat{\mathrm{Err}}_k,
\end{equation}
\[
\widehat{\tau}_0^2=\frac{1}{K(K+1)}\sum_{0\le i<j\le K}\Bigl\{\bigl(\widehat{\mathrm{Err}}_i-\widehat{\mathrm{Err}}_j\bigr)^2-\widehat{\sigma}_i^2-\widehat{\sigma}_j^2+2\widehat{\sigma}_{i,j}\Bigr\},
\]
respectively, then using the normal-normal update \cite{efron1996empirical}
\[
\mathrm{Err}_0\mid \widehat{\mathrm{Err}}_0\;\sim\;\mathcal{N}\!\left(\frac{\widehat{\sigma}_0^{-2}\widehat{\mathrm{Err}}_0+\widehat{\tau}_0^{-2}\widehat{\mu}_0}{\widehat{\sigma}_0^{-2}+\widehat{\tau}_0^{-2}},\;\frac{1}{\widehat{\sigma}_0^{-2}+\widehat{\tau}_0^{-2}}\right).
\]
In the end, the EB point estimator is
\begin{equation}
\label{eq:EBest}
\widehat{\mathrm{Err}}_0^{EB}=\frac{\widehat{\sigma}_0^{-2}\widehat{\mathrm{Err}}_0+\widehat{\tau}_0^{-2}\widehat{\mu}_0}{\widehat{\sigma}_0^{-2}+\widehat{\tau}_0^{-2}}.
\end{equation}
Note that $\widehat{\mu}_0$ in \eqref{eq:momentest} is the usual random-split cross-validation estimator used in practice to estimate average model performance. 
\begin{remark}
It is possible that the simple estimator $\widehat{\tau}_0^2$ takes a negative value. In such a case, one may either increase the number of cross-validation splits, $K,$ and updates the moment estimator for $\tau_0^2$, or simply let   $\widehat{\tau}_0=0$ and $\widehat{\mathrm{Err}_0}^{EB}=\hat{\mu}_0.$ 
\end{remark}
\subsection{Applications}
Assume observed data are $\mathbf{X}=\{X_i=(Z_i,Y_i): i=1,\ldots,n\}$ with $n$ i.i.d observations, where $Z$ is covariate vector and $Y$ is outcome.  Let $\widehat{Y}_k(Z)$ denote the prediction from $\widehat{\mathcal{M}}_{\mathbf{D}_1^{(k)}}$.

\paragraph{Continuous outcome.} For continuous $Y$, a common loss is the mean squared prediction error (MSPE):
\begin{equation}
\label{eq:mseest}
\widehat{\mathrm{Err}}_k=\frac{1}{n_2}\sum_{(Z_i,Y_i)\in\mathbf{D}_2^{(k)}}\bigl\{Y_i-\widehat{Y}_k(Z_i)\bigr\}^2,\quad k=0,1,\ldots,K.
\end{equation}
An estimator of $\mathrm{Cov}(\widehat{\mathrm{Err}}_k,\widehat{\mathrm{Err}}_\ell)$ is
\begin{equation}
\label{eq:msecovest}
\widehat{{\sigma}}_{k,\ell}=\frac{1}{n_2^2}\sum_{(Z_i,Y_i)\in\mathbf{D}_2^{(k)}\cap\mathbf{D}_2^{(\ell)}}\Bigl[\{Y_i-\widehat{Y}_k(Z_i)\}^2-\widehat{\mathrm{Err}}_k\Bigr]\Bigl[\{Y_i-\widehat{Y}_\ell(Z_i)\}^2-\widehat{\mathrm{Err}}_\ell\Bigr].
\end{equation}
With $\{\widehat{\mathrm{Err}}_k\}_{k=0}^K$ and $\widehat{\boldsymbol{\Sigma}},$ our goal is to estimate the MSPE $\mathrm{Err}_0=\mathbb{E}[\{Y_0-\widehat{Y}_0(Z_0)\}^2]$.

\paragraph{Binary outcome.} For binary $Y$ and continuous risk score $\widehat{Y}_k(Z)$, we use the c-index (the area under the receiver operating characteristic curve):
\begin{equation}
\label{eq:aucest}
\widehat{\mathrm{Err}}_k=\frac{1}{m_1^{(k)}m_0^{(k)}}\sum_{(Z_i,Y_i)\in\mathbf{D}_2^{(k)0}}\sum_{(Z_j,Y_j)\in\mathbf{D}_2^{(k)1}} \widehat{U}_{ij}^{(k)}
\end{equation}
to estimate the predictive performance \cite{pepe2003statistical, zhao2011auc}, where $\widehat{U}_{ij}^{(k)}=\mathbf{1}\{\widehat{Y}_k(Z_i)<\widehat{Y}_k(Z_j)\}$, $\mathbf{D}_2^{(k)j}=\{(Z_i,Y_i)\in\mathbf{D}_2^{(k)}:Y_i=j\}$, and $m_j^{(k)}=\#\mathbf{D}_2^{(k)j}$ for $j\in\{0,1\}$. The covariance between $\widehat{\mathrm{Err}}_k$ and $\widehat{\mathrm{Err}}_\ell$ can be estimated by the U-statistic 
{\small
\begin{align}
\widehat{\sigma}_{k,\ell}=&\;\frac{1}{m_0^{(k)}m_0^{(\ell)}}\sum_{(Z_i,Y_i)\in\mathbf{D}_2^{(k)0}\cap\mathbf{D}_2^{(\ell)0}}\Bigl\{\frac{1}{m_1^{(k)}}\sum_{(Z_j,Y_j)\in\mathbf{D}_2^{(k)1}}\widehat{U}_{ij}^{(k)}-\widehat{\mathrm{Err}}_k\Bigr\}\Bigl\{\frac{1}{m_1^{(\ell)}}\sum_{(Z_j,Y_j)\in\mathbf{D}_2^{(\ell)1}}\widehat{U}_{ij}^{(\ell)}-\widehat{\mathrm{Err}}_\ell\Bigr\} \nonumber\\
&+\frac{1}{m_1^{(k)}m_1^{(\ell)}}\sum_{(Z_i,Y_i)\in\mathbf{D}_2^{(k)1}\cap\mathbf{D}_2^{(\ell)1}}\Bigl\{\frac{1}{m_0^{(k)}}\sum_{(Z_j,Y_j)\in\mathbf{D}_2^{(k)0}}\widehat{U}_{ji}^{(k)}-\widehat{\mathrm{Err}}_k\Bigr\}\Bigl\{\frac{1}{m_0^{(\ell)}}\sum_{(Z_j,Y_j)\in\mathbf{D}_2^{(\ell)0}}\widehat{U}_{ji}^{(\ell)}-\widehat{\mathrm{Err}}_\ell\Bigr\}. \label{eq:auccovest}
\end{align}}
With $\{\widehat{\mathrm{Err}}_k\}_{k=0}^K$ and $\widehat{\boldsymbol{\Sigma}}$, our objective is to estimate $\mathrm{Err}_0=\mathbb{P}\bigl\{\widehat{Y}_0(Z_1)<\widehat{Y}_0(Z_2)\mid Y_1=0, Y_2=1\bigr\}.$

\section{Simulation Study} \label{sec:simulation}
In this section, we perform two sets of simulations to examine the performance of our proposal.
\subsection{Continuous Outcome} \label{sec:sim1}
\begin{figure}[ht]
    \centering
    \begin{minipage}{0.49\textwidth}
        \centering
        \includegraphics[width=\linewidth]{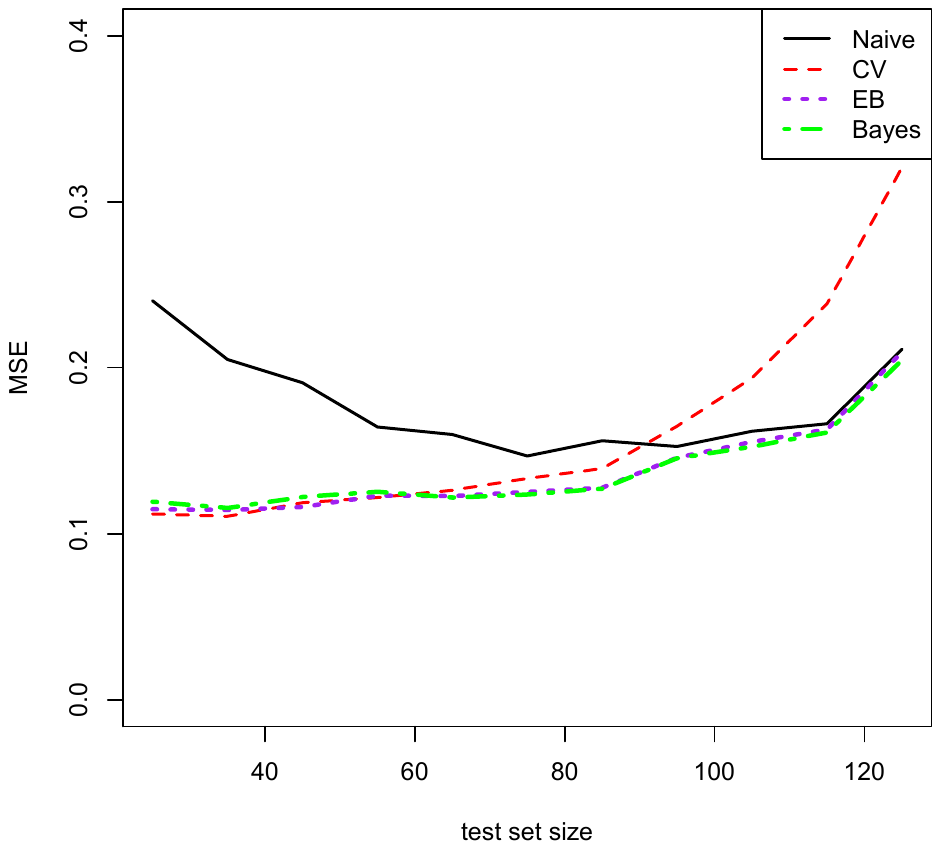}
        \caption{Mean absolute estimation error of $\widehat{\mathrm{Err}}_0$, $\widehat{\mathrm{Err}}_0^{CV}$, $\widehat{\mathrm{Err}}_0^{EB}$, and $\widehat{\mathrm{Err}}_0^{B}$ for varying test set sizes. Here $\mathrm{Err}_0$ is the mean squared prediction error for a continuous outcome.}
        \label{fig:simmse}
    \end{minipage}\hfill
    \begin{minipage}{0.49\textwidth}
        \centering
        \includegraphics[width=\linewidth]{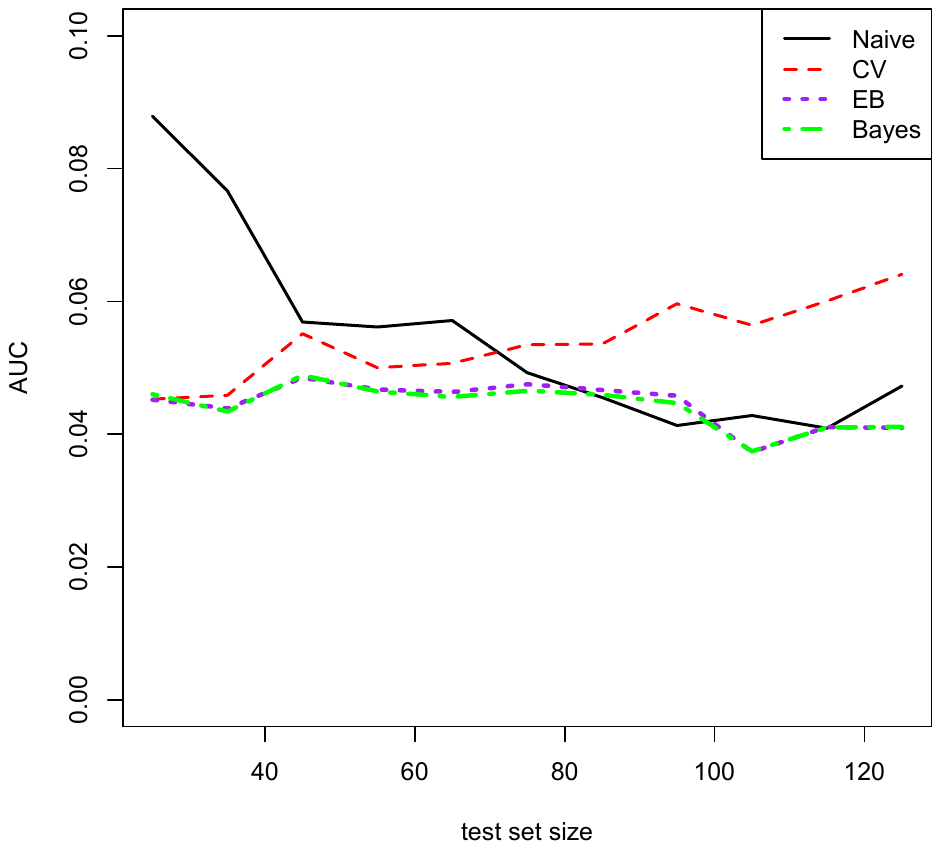}
        \caption{Mean absolute estimation error of $\widehat{\mathrm{Err}}_0$, $\widehat{\mathrm{Err}}_0^{CV}$, $\widehat{\mathrm{Err}}_0^{EB}$, and $\widehat{\mathrm{Err}}_0^{B}$ for varying test set sizes. Here $\mathrm{Err}_0$ is the area under the Receiver's operating characteristics curve (AUC) for a binary outcome.}
        \label{fig:simbinary}
    \end{minipage}
\end{figure}

We generate $n=150$ independent observations $Z_i=(Z_{i1}, \cdots, Z_{ip})'\sim\mathcal{N}(0,\mathbf{I}_{50})$ and
\[
Y_i=0.5(Z_{i1}+Z_{i2}+Z_{i3}+Z_{i4})+\varepsilon_i,\qquad \varepsilon_i\sim\mathcal{N}(0,1).
\]
We then split $\mathbf{X}=\mathbf{D}_1^{(0)}\cup\mathbf{D}_2^{(0)}$ with sizes $n_1$ and $n_2=n-n_1$ and fit a lasso regularized linear model on $\mathbf{D}_1^{(0)}$ (the lasso penalty parameter fixed at $\lambda=0.10$) \cite{tibshirani1996regression}. Denote the fitted intercept and slope by $\widehat{\alpha}_0$ and $\widehat{\beta}_0$, giving the prediction rule $\widehat{Y}_0(Z)=\widehat{\alpha}_0+\widehat{\beta}_0'Z$. The true MSPE is
\[
\mathrm{Err}_0=\widehat{\alpha}_0^2+\lVert\widehat{\beta}_0-\beta_0\rVert_2^2+1,\qquad \beta_0=(0.5,0.5,0.5,0.5,0,\ldots,0)'.
\]
The naive estimator of $\mathrm{Err}_0$ is
\[
\widehat{\mathrm{Err}}_0=\frac{1}{n_2}\sum_{(Z_i,Y_i)\in\mathbf{D}_2^{(0)}}\bigl\{Y_i-\widehat{Y}_0(Z_i)\bigr\}^2.
\]
For $k=1,\ldots,K=39,$ we repeat random splits of $\mathbf{X}$ into $(\mathbf{D}_1^{(k)},\mathbf{D}_2^{(k)})$, fit lasso regularized linear regression, compute $\widehat{\mathrm{Err}}_k$ by \eqref{eq:mseest}, and estimate covariances by \eqref{eq:msecovest} to form $\widehat{\boldsymbol{\Sigma}}$. In the end, we compute four estimates for $\mathrm{Err}_0:$
\begin{enumerate}
  \item Naive estimate $\widehat{\mathrm{Err}}_0$;
  \item Cross-validation (random-split) estimate $\widehat{\mathrm{Err}}_0^{CV}=\widehat{\mu}_0$ via \eqref{eq:momentest};
  \item Bayesian estimate $\widehat{\mathrm{Err}}_0^{B}$ via \eqref{eq:Best};
  \item Empirical Bayes estimate $\widehat{\mathrm{Err}}_0^{EB}$ via \eqref{eq:EBest}.
\end{enumerate}
We repeat the aforementioned data generation and parameter estimation 299 times. Figure~\ref{fig:simmse} summarizes the mean absolute estimation error of four estimates for different combinations of $(n_1, n_2)$. As expected, when $n_2$ is small, $\widehat{\mathrm{Err}}_0^{CV}$ outperforms the naive $\widehat{\mathrm{Err}}_0$. When $n_2$ is large (so $n_1$ is small), models trained on $\mathbf{D}_1^{(k)}$ differ substantially; $\widehat{\mathrm{Err}}_0^{CV}$ can be biased for estimating $\mathrm{Err}_0$ and be inferior to the naive estimator. The proposed $\widehat{\mathrm{Err}}_0^{B}$ and $\widehat{\mathrm{Err}}_0^{EB}$ have a very similar accuracy and are consistently no worse than the best of $\widehat{\mathrm{Err}}_0$ and $\widehat{\mathrm{Err}}_0^{CV}$.  In addition, we examine the empirical coverage levels of 95\% credible intervals for $\mathrm{Err}_0$, which vary from 89\% to 95\% across different $n_2.$ 

\subsubsection{Binary Outcomes}
Let $Y_i\in\{0,1\}$ be generated by
\[
Y_i=\mathbf{1}\bigl\{0.5(Z_{i1}+Z_{i2}+Z_{i3}+Z_{i4})+\varepsilon_i>0\bigr\},\qquad \varepsilon_i\sim\mathcal{N}(0,1),
\]
where $\mathbf{1}(\cdot)$ is an indicator function. We fit a lasso regularized logistic regression model on $\mathbf{D}_1^{(k)}$ (the penalty parameter $\lambda=0.13$), yielding a ``risk'' score $\widehat{Y}_k(Z)=\widehat{\alpha}_k+\widehat{\beta}_k'Z$ and estimated probability $\widehat{\Pr}(Y=1\mid Z)=\mbox{expit}\{\widehat{Y}_k(Z)\}$. Then, the c-index in the test set is estimated via \eqref{eq:aucest}. The variance-covariance matrix of of $K+1$ estimated c-indeces, $\widehat{\boldsymbol{\Sigma}},$ is calculated via \eqref{eq:auccovest}. Lastly, these four estimates for $\mathrm{Err}_0$ described in section \ref{sec:sim1} are obtained. After 299 simulations, we observe that the same qualitative pattern holds: when $n_2$ is small, $\widehat{\mathrm{Err}}_0^{CV}$ is better than the naive estimator; when $n_2$ is large, the naive estimator is better; and the proposed $\widehat{\mathrm{Err}}_0^{B}$ and $\widehat{\mathrm{Err}}_0^{EB}$ match or improve upon the best of the two baseline estimates. The detailed results are summarized in Figure \ref{fig:simbinary}. The empirical coverage levels of these 95\% credible intervals for c-index are slightly lower than those in section \ref{sec:sim1}: they vary from 85\% to 93\% for different $n_2.$ 

\section{Example}
We use \emph{Bike Sharing} data from the UCI Machine Learning Repository (\url{https://archive.ics.uci.edu/dataset/275/bike+sharing+dataset}) \cite{fanaee2014event}. The hourly dataset contains 17{,}379 records and (reported) 12 variables, recording bike-sharing activity over two years. Variables include date, year (2011--2012), month (1--12), season (winter, spring, summer, fall), hour (0--23), holiday (yes/no), day of week, working-day indicator, weather (clear, few clouds, partly cloudy, mist + cloudy, mist + broken clouds, mist + few clouds, mist, light snow, light rain + thunderstorm + scattered clouds, light rain + scattered clouds, heavy rain + ice pellets + thunderstorm + mist, snow + fog), normalized temperature, normalized ``feels-like'' temperature, normalized humidity, and normalized wind speed. We use these features to predict hourly total rental counts and evaluate the resulting prediction model.

\begin{figure}[ht]
    \centering
    \begin{minipage}{0.49\textwidth}
    \centering
    \includegraphics[width=\linewidth]{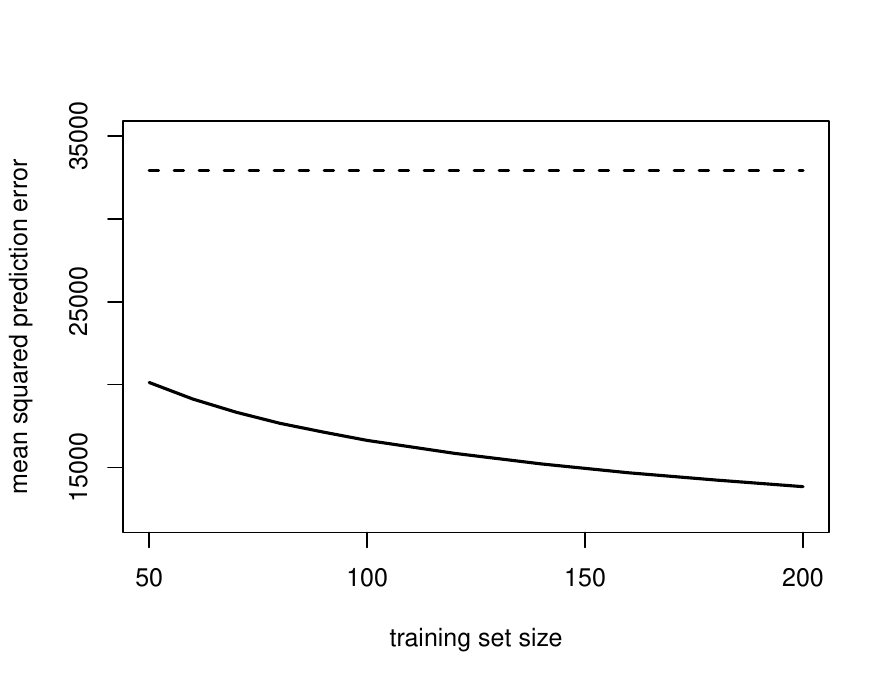}
    \caption{Mean squared prediction error of a random forest for hourly rental counts with the sample size in training sets varying from 50 to 200. The dotted line is the mean squared prediction error of predicting the hourly counts by their sample mean, a constant.}
    \label{fig:mseexample}
    \end{minipage}\hfill
    \begin{minipage}{0.49\textwidth}
    \centering
    \includegraphics[width=\linewidth]{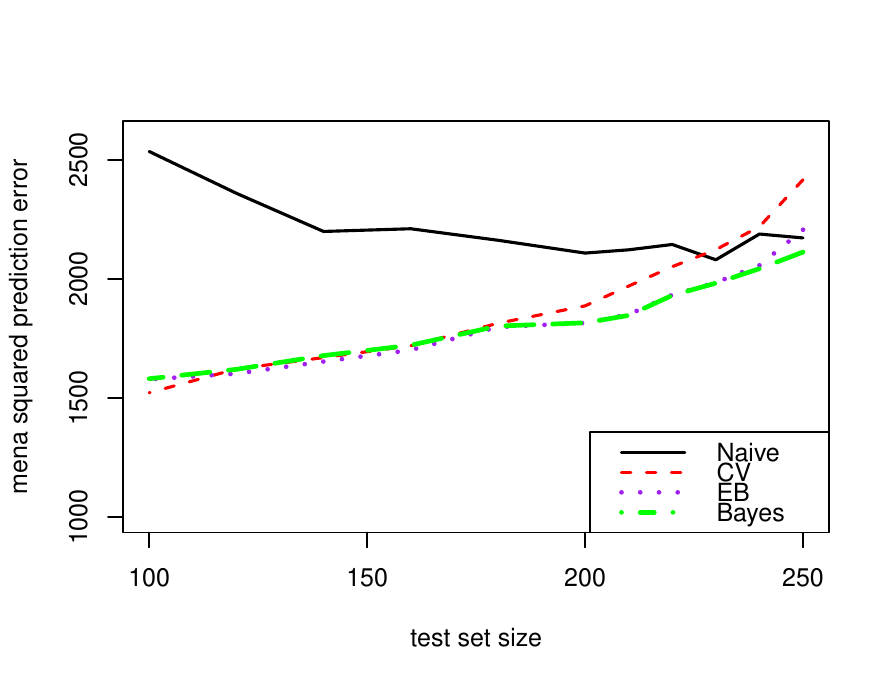}
    \caption{Mean absolute estimation error of $\widehat{\mathrm{Err}}_0$, $\widehat{\mathrm{Err}}_0^{CV}$, $\widehat{\mathrm{Err}}_0^{EB}$, and $\widehat{\mathrm{Err}}_0^{B}$ for varying validation sizes in the bike-sharing data. Here $\mathrm{Err}_0$ is the mean squared prediction error of a random forest for hourly rental counts.}
    \label{fig:cvexample}
    \end{minipage}
\end{figure}

We first split the dataset into two parts with 300 observations in the first part and 17{,}079 in the second. The second part is reserved to estimate the performance of the model developed from the first part; owing to its large size, we treat this estimate as the ``true'' $\mathrm{Err}_0$ for benchmarking. Within the first part, we repeatedly split data into a training set of size $n_1$ and a validation set of size $n_2=300-n_1$. A random forest (500 trees) is trained on the training set to obtain $\widehat{\mathcal{M}}_{\mathbf{D}_1^{(0)}}$. The validation set yields the naive estimate of MSPE, $\widehat{\mathrm{Err}}_0$. We then perform $K=40$ additional random splits for the first part of data to obtain $\{(\widehat{\mathcal{M}}_{\mathbf{D}_1^{(k)}},\widehat{\mathrm{Err}}_k):k=1,\ldots,40\}$, and compute $\widehat{\mathrm{Err}}_0^{CV}$, $\widehat{\mathrm{Err}}_0^{B}$, and $\widehat{\mathrm{Err}}_0^{EB}$. Using the MSPE on the large second part as the truth, we compute the mean absolute estimation error for each method. Repeating this process 299 times, the true MSPE and the mean absolute estimation error of its estimator for different combinations of $(n_1,n_2)$ are summarized in Figures~ \ref{fig:mseexample} and \ref{fig:cvexample}, respectively.

Consistent with the simulation study, when the \emph{training set is large and the validation set is small}, $\widehat{\mathrm{Err}}_0^{CV}$ substantially outperforms the naive estimator $\widehat{\mathrm{Err}}_0$, and both $\widehat{\mathrm{Err}}_0^{B}$ and $\widehat{\mathrm{Err}}_0^{EB}$ perform comparably to $\widehat{\mathrm{Err}}_0^{CV}$. Conversely, when the \emph{training set is small and the validation set is large}, the naive estimator $\widehat{\mathrm{Err}}_0$ tends to outperform $\widehat{\mathrm{Err}}_0^{CV}$, while the proposed estimators perform comparably to the naive one. For certain $(n_1, n_2)$ configurations, the proposed estimators achieve higher accuracy than both baseline estimators. For example, when $n_1=80$ and $n_2=220$, the mean absolute estimation errors of $\widehat{\mathrm{Err}}_0$ and $\widehat{\mathrm{Err}}_0^{CV}$ are 2050 and 2040, respectively, whereas those of $\widehat{\mathrm{Err}}_0^{B}$ and $\widehat{\mathrm{Err}}_0^{EB}$ are 1925 and 1940. The true MSPE in this setting is 17,665, implying relative estimation errors ranging from 10.9\% to 11.6\%. As a reference, the MSPE of a naive prediction rule that predicts hourly rides by a constant (the sample mean) is 32,901, suggesting that the average $R^2$ of random forest models trained on a training set of 80 observations is approximately 46.3\%. 

Finally, we examine the empirical coverage of posterior-quantile intervals for $\mathrm{Err}_0$. The empirical coverage levels are 95\%, 93\%, 95\%, and 96\% for $n_1=50, 100, 140,$ and $200$, respectively, suggesting that simple credible intervals may serve as useful uncertainty measures for $\widehat{\mathrm{Err}}_0^{B}$ in this example.

\section{Discussion}
We propose a new approach for estimating the performance of a prediction model trained on a selected subset of observed data. Our method enables researchers to present both a prediction model and a corresponding estimate of its predictive performance to potential users. The proposed estimators adaptively combine information from a naïve single-split estimate and conventional cross-validation, achieving improved accuracy across a wide range of settings.

Furthermore, our numerical results indicate that the conventional cross-validation estimator performs particularly well when the training set is larger than the testing set -- a situation common in practice -- even though it is typically viewed as a tool for evaluating modeling strategies rather than specific prediction models.

Several extensions to our proposed approach are possible. First, one may relax the Gaussian assumption for random effects, $\{\mathrm{Err}_k\}_{k=0}^K,$ and estimate this mixing distribution nonparametrically, when $K$ is large. Second, constructing a valid frequentist confidence interval for $\mathrm{Err}_0$ is of interest. Our simulation study shows that the frequentist coverage level of credible intervals based on the posterior distribution could be slightly lower than the nominal level. Nonparametric calibration via bootstrap may be a feasible approach to correct this under coverage \cite{chen2025empirical}.


\begin{figure}[ht]
   
\end{figure}

\end{document}